# Adaptive Local Structure Consistency based Heterogeneous Remote Sensing Change Detection

Lin Lei†, Yuli Sun†, and Gangyao Kuang, *Senior Member, IEEE*

*Abstract*—Change detection of heterogeneous remote sensing images is a challenging topic, which plays an important role in natural disaster emergency response. Due to the different imaging mechanisms of heterogeneous sensors, it is hard to directly compare the images. To address this challenge, we explore an unsupervised change detection method based on adaptive local structure consistency (ALSC) between heterogeneous images in this letter, which constructs an adaptive graph representing the local structure for each patch in one image domain and then projects this graph to the other image domain to measure the change level. This local structure consistency exploits the fact that the heterogeneous images share the same structure information for the same ground object, which is imaging modality-invariant. To avoid the heterogeneous data confusion, the pixelwise change image is calculated in the same image domain by graph projection. By comparing with some state-of-the-art methods, the experimental results show the effectiveness of the proposed ALSC based change detection method.

## I. INTRODUCTION

Change detection (CD) is a process of identifying changes of objects or phenomenon by analyzing the multitemporal remote sensing images acquired over the same geographical area[1], [2]. Recently, heterogeneous CD has become an interesting topic that allows the remote sensing community to make full use of the wide range of available earth observation satellites by considering the combined use of heterogeneous remote sensing images. In particular, in the immediate assessment of emergency disasters (such as earthquake or flood), heterogeneous CD plays an important role. In this case, the pre-event synthetic aperture radar (SAR) image is sometimes unavailable and the qualified post-event optical image cannot be obtained due to the adverse atmospheric conditions.

Heterogeneous CD is a challenging task since the sensors measure different physical quantities and show quite different characteristics for the same object in the images. Therefore, different from the algebraic methods (such as difference method, ratio method and log-ratio method) used in the homogeneous CD, it is impossible to compare the heterogeneous images directly to calculate the difference image (DI). According to the different generation processes of change map (CM), the existing methods of heterogeneous CD can be roughly divided into three categories [3]: 1) classification-based methods, such as post-classification comparison (PCC) [4], multitemporal segmentation and compound classification based method (MSCC) [5], [6]; 2) deep-learning based methods, such as symmetric convolutional coupling network (SCCN) [7], logarithmic transformation feature learning based method (LTFL) [8], and conditional generative adversarial network (cGAN) [9]; 3) traditional DI-based methods, such as the local joint distributions with manifold learning based method [10], Markov model for multimodal change detection (M3CD) [11], supervised homogeneous pixel transformation (HPT) method [12], and affinity matrix based image regression (AM-IR) method [13]. The goal of the heterogeneous CD approach is to transform the "incomparable" images into a common space to make them "comparable", such as the category space, the learned high-dimensional or the constructed feature space[3].

Recently, the self-similarity property is exploited to project the pre-event (or post-event) image to the post-event (or pre-event) image and then obtain a pixelwise DI for heterogeneous CD. In [14], the self-similarity property is used to complete the fractal projection, which contains a fractal encoding step and a fractal projection/decoding step. In [3], the self-similarity is used to complete self-expression, which learns a patch similarity graph matrix (PSGM) for each image and then measures the change level by using this PSGM. In this letter, the self-similarity is further used to construct graphs representing the local structures for each image and establish the relationships between heterogeneous images, which is similar to the patch similarly graph based method in [15]. However, instead of using a fixed graph as [15], the proposed method adaptively learns a distance induced probabilistic graph, which is more robust. This local structure consistency exploits the fact that the heterogeneous images share the same structure information for the same ground object, which is imaging modality-invariant. Different from the previous fractal projection and PSGM based methods, the proposed ALSC based method neither reconstructs the image nor transforms the image to the other domain. It only focuses on the changes of local structure. At the same time, it also takes into account the prior statistics of heterogeneous images that are not used in fractal projection [14] and PSGM [3], which makes the CD results more accurate. The contributions of this letter are as follows.

1) A heterogeneous CD method based on adaptive local structure consistency is proposed, which adaptively constructs the local structure for each image, and then measures the consistency between local structure of two images.

2) The structure difference is calculated within the same image domain by the graph projection, which avoids the

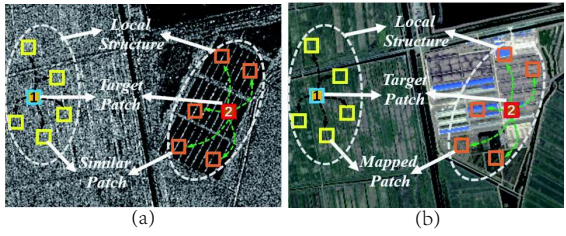

Fig. 1. Illustration of local structure consistency in heterogeneous images: (a) SAR image; (b) Optical image. The local structure of the unchanged target patch 1 in the SAR image can be preserved by the patch 1 in the optical image. However, for the changed target patch 2, the local structure in the SAR image is no longer conformed by the optical image.

heterogeneous data confusion.

3) The ALSC based method is completely unsupervised, and it exploits the inherent structure property of any satellite image that appeals quite imaging modality-invariant, so it has great flexibility to deal with various heterogeneous image processing tasks.

## II. ALSC BASED HETEROGENEOUS CD METHOD

We consider two coregistered images acquired by heterogeneous sensors at different times, which are denoted as $\mathbf{X} = \{x(m,n,c) | 1 \le m \le M, 1 \le n \le N, 1 \le c \le C_{\mathbf{X}}\}$ and $\mathbf{Y} = \{y(m,n,c) | 1 \le m \le M, 1 \le n \le N, 1 \le c \le C_{\mathbf{Y}}\}$ in $\mathcal{X}$ domain and $\mathcal{Y}$ domain, respectively. Here, $M$, $N$ and $C_{\mathbf{X}}$ ($C_{\mathbf{Y}}$) represent the height, width and channel number of the image, respectively.

With the self-similarity property, for each small patch in the image, there are always some other patches very similar to it in an extended (local) window centered on it. The relationship between this target patch and its similar patches (not all the patches) in an extended window (not the whole image) can be regarded as the "local structure" of this target patch, which is quite well preserved across the different types of imaging modality [15]. As illustrated in Fig. 1, in the SAR image $\mathbf{X}$, the local structure of target patch $\mathbf{X}_i$ is represented by the relationship between $\mathbf{X}_i$ and its similar patches $\mathbf{X}_j$. If the area represented by these patches has not changed in the event, the local structure of target patch $\mathbf{X}_i$ can be preserved by the patch $\mathbf{Y}_i$ in the optical image $\mathbf{Y}$, which shows that the patches $\mathbf{Y}_j$ and the patch $\mathbf{Y}_i$ are also very similar. On the contrary, if the area represented by $\mathbf{X}_i$ has changed in the event, the local structure is no longer preserved by $\mathbf{Y}_i$, showing that $\mathbf{Y}_i$ and $\mathbf{Y}_j$ are very different. Therefore, we can see that the ALSC based heterogeneous CD method mainly needs to solve two problems: how to construct local structure and how to measure structural difference. The ALSC based method consists of four steps: 1) constructing adaptive local structure; 2) calculating the structure difference; 3) generating the DI; 4) obtaining the binary CM.

### A. Adaptive local structure

For a square target patch, $\mathbf{X}_i = \{x(m_i + \vartheta_m, n_i + \vartheta_n, c) | \vartheta_m, \vartheta_n \in [-p, p], 1 \le c \le C_{\mathbf{X}}\} \in \mathbb{R}^{(2p+1) \times (2p+1) \times C_{\mathbf{X}}}$ centered on position $(m_i, n_i)$, in order to capture the local structure of $\mathbf{X}_i$, we construct a graph by adaptively assigning a probability $S_{i,j}^{\mathbf{X}}$ for $\mathbf{X}_j$ as the neighborhood of $\mathbf{X}_i$, and the probability is treated as the similarity between two patches, where $\mathbf{X}_j$ is another patch centered on $(m_j, n_j)$ in a $\omega \times \omega$ search window $\mathbb{W}$ centered on $\mathbf{X}_i$ and $(m_j, n_j) \ne (m_i, n_i)$. Here, a search step is used to make the position distance between adjacent patches in $\mathbb{W}$ equal to $\Delta_s$, which can reduce the number of patches and avoid over aggregation of patches.

Usually, a smaller distance between $\mathbf{X}_i$ and $\mathbf{X}_j$ should be assigned a larger probability $S_{i,j}^{\mathbf{X}}$. To achieve this goal, we can solve the following problem

$$\min_{S_{i,j}^{\mathbf{X}}} \sum_{j=1}^{N_s} dist_{i,j}^{\mathcal{X}} S_{i,j}^{\mathbf{X}} + \gamma \left(S_{i,j}^{\mathbf{X}}\right)^2 s.t.\ 0 \le S_{i,j}^{\mathbf{X}} \le 1,\ \sum_{j=1}^{N_s} S_{i,j}^{\mathbf{X}} = 1 \quad (1)$$

where $dist_{i,j}^{\mathcal{X}}$ represents a distance measure of two patches $\mathbf{X}_i$ and $\mathbf{X}_j$, $N_s$ is the the number of all potential neighbors of $\mathbf{X}_i$ in the search window $\mathbb{W}$, and $\gamma > 0$ is the trade-off parameter. The second term of the objective function in (1) is a regularization, which is used to avoid getting the trivial solutions, i.e., only the nearest patch can be the neighbor of $\mathbf{X}_i$ with probability 1 (when $\gamma = 0$). If we only focus on the second term (when $\gamma \to \infty$), the optimal solution of (1) is that all the patches in $\mathbb{W}$ can be the neighbors of $\mathbf{X}_i$ with the same probability $1/N_s$.

The patch distance $dist_{i,j}^{\mathcal{X}}$ should be computed based on the prior statistics of the image $\mathbf{X}$. For two patches $\mathbf{X}_i$ and $\mathbf{X}_j$, we vectorized them and denote each element as $x_i(q)$ and $x_i(q)$ with $x_j(q)$, respectively. 1) For the optical image, the traditional Euclidean distance is usually chosen

$$dist_{i,j}^{\mathcal{X}} = \|\mathbf{X}_i - \mathbf{X}_j\|_F^2 \quad (2)$$

2) For the SAR image, with the multiplicative speckle noise modeled by a Gamma distribution, the following distance criterion [16] can be used

$$dist_{i,j}^{\mathcal{X}} = \sum_{q=1}^{(2p+1)^2 C_{\mathbf{X}}} \log \left( \frac{x_i(q) + x_j(q)}{2\sqrt{x_i(q) x_j(q)}} \right) \quad (3)$$

It should be noted that the distance calculation function is not fixed, which should be determined according to the prior statistical information of the image.

Denote $S_i^{\mathbf{X}} \in \mathbb{R}^{N_s}$ and $dist_i^{\mathcal{X}} \in \mathbb{R}^{N_s}$ as column vectors composed of $S_{i,j}^{\mathbf{X}}$ and $dist_{i,j}^{\mathcal{X}}$ respectively, and denote $\mathbf{1}$ as a column vector with all the elements being one, the problem (1) can be reformulated as

$$\min_{(S_i^{\mathbf{X}})^T \mathbf{1} = 1, 0 \le S_{i,j}^{\mathbf{X}} \le 1} \left\| S_i^{\mathbf{X}} + \frac{1}{2\gamma} dist_i^{\mathcal{X}} \right\|_2^2 \quad (4)$$

This problem naturally has a sparse solution, which means that the target patch $\mathbf{X}_i$ is only connected with its $K$-nearest neighbors (K-NN). That is, $S_i^{\mathbf{X}}$ has only $K$ nonzero entries with $1 \le K \le N_s - 1$. This problem can be solved with a closed solution as in [17]. By denoting $dist_{i,i_{\mathbf{X}}^{(h)}}^{\mathcal{X}}$ with $h = 1, 2, \cdots, N_s$ as the $h$-th smallest element of $dist_i^{\mathcal{X}}$, we give a summary of this efficient solution as shown in Algorithm 1 of Table I, and more details can be found in [17].

TABLE I
IMPLEMENTATION STEPS OF ALGORITHM 1.

| Algorithm 1. The optimization algorithm of minimization (4) |
|---|
| **Input:** Distance vector $dist_i^{\mathcal{X}}$ and the number of nearest neighbors $K$. |
| 1. Sorting $dist_i^{\mathcal{X}}$ in ascending order as $dist_{i,i_{\mathbf{X}}^{(1)}}^{\mathcal{X}}, \cdots, dist_{i,i_{\mathbf{X}}^{(N_s)}}^{\mathcal{X}}$. |
| 2. Calculating the $S_i^{\mathbf{X}}$ with $$S_{i,i_{\mathbf{X}}^{(h)}}^{\mathbf{X}} = \begin{cases} \frac{dist_{i,i_{\mathbf{X}}^{(K+1)}}^{\mathcal{X}} - dist_{i,i_{\mathbf{X}}^{(h)}}^{\mathcal{X}}}{K dist_{i,i_{\mathbf{X}}^{(K+1)}}^{\mathcal{X}} - \sum_{r=1}^{K} dist_{i,i_{\mathbf{X}}^{(r)}}^{\mathcal{X}}}, & h \leq K \\ 0, & h > K \end{cases}$$ |
| **Output:** The similarity vector $S_i^{\mathbf{X}}$. |

From Algorithm 1, we can find that the regularization parameter $\gamma$ is replaced by the number of neighbors $K$, which are equivalent when we set $\gamma$ to be

$$\gamma = \frac{K}{2} dist_{i,i_{\mathbf{X}}^{(K+1)}}^{\mathcal{X}} - \frac{1}{2} \sum_{r=1}^{K} dist_{i,i_{\mathbf{X}}^{(r)}}^{\mathcal{X}} \quad (5)$$

In this way, the search of parameter $\gamma$ can be better handled by searching the neighborhood size $K$, which is more intuitive (it has explicit meaning) and easy to tune (it is an integer).

*B. Calculate the structure difference*

As the local structure of patches $\mathbf{X}_i$ and $\mathbf{Y}_i$ can be represented by $S_i^{\mathbf{X}}$ and $S_i^{\mathbf{Y}}$, we need to compare them to calculate the structure difference. However, because $S_i^{\mathbf{X}}$ and $S_i^{\mathbf{Y}}$ are generated based on different distance vectors $dist_i^{\mathcal{X}}$ and $dist_i^{\mathcal{Y}}$, which are calculated on different domains $\mathcal{X}$ and $\mathcal{Y}$, directly comparing $S_i^{\mathbf{X}}$ and $S_i^{\mathbf{Y}}$ (such as $\|S_i^{\mathbf{X}} - S_i^{\mathbf{Y}}\|_2$) will cause the confusion of heterogeneous data. To avoid the confusion, we first project the graph $S_i^{\mathbf{X}}$ (or $S_i^{\mathbf{Y}}$) of image domain $\mathcal{X}$ (or $\mathcal{Y}$) to the other image domain $\mathcal{Y}$ (or $\mathcal{X}$), and then compare the difference in the same image domain to obtain the structure difference. Therefore, the forward structure difference $f_i^{\mathbf{Y}}$ and the backward structure difference $f_i^{\mathbf{X}}$ can be computed as

$$f_i^{\mathbf{Y}} = \sum_{h=1}^{h=K} dist_{i_{\mathbf{Y}}^{(h)}, i_{\mathbf{X}}^{(h)}}^{\mathcal{Y}} S_{i,i_{\mathbf{X}}^{(h)}}^{\mathbf{X}}, \quad f_i^{\mathbf{X}} = \sum_{h=1}^{h=K} dist_{i_{\mathbf{Y}}^{(h)}, i_{\mathbf{X}}^{(h)}}^{\mathcal{X}} S_{i,i_{\mathbf{Y}}^{(h)}}^{\mathbf{Y}} \quad (6)$$

The intuitive explanation of $f_i^{\mathbf{Y}}$ in (6) is: for the patch $\mathbf{X}_j$ with $j = i_{\mathbf{X}}^{(h)}$ in $S_i^{\mathbf{X}}$, which is the $h$-th nearest neighbor of target patch $\mathbf{X}_i$, the $dist_{i_{\mathbf{Y}}^{(h)}, i_{\mathbf{X}}^{(h)}}^{\mathcal{Y}}$ is the patch difference between the projected patch $\mathbf{Y}_j$ (that is $\mathbf{Y}_{i_{\mathbf{X}}^{(h)}}$) and the $h$-th nearest neighbor of patch $\mathbf{Y}_i$ (that is $\mathbf{Y}_{i_{\mathbf{Y}}^{(h)}}$), and the $S_{i,i_{\mathbf{X}}^{(h)}}^{\mathbf{X}}$ is the weight of this patch difference. Therefore, we can find that $S_i^{\mathbf{X}}$ is projected to the $\mathcal{Y}$ domain by transmitting the position information ($i_{\mathbf{X}}^{(h)}$ in the $dist_{i_{\mathbf{Y}}^{(h)}, i_{\mathbf{X}}^{(h)}}^{\mathcal{Y}}$) and probability information ($S_{i,i_{\mathbf{X}}^{(h)}}^{\mathbf{X}}$), respectively. Then, the patch difference is calculated in the same image domain, which can avoid the heterogeneous data confusion.

*C. Generate the difference image*

Once the forward structure difference $f_i^{\mathbf{Y}}$ is calculated, it is assigned to all the pixels of patch $\mathbf{Y}_i$. After we calculate the forward structure differences for all the overlapping patches of the two images, for each pixel $j$, $j = 1, \cdots, MN$ in the forward DI, there is a set $F_j^{\mathbf{Y}}$ of structure difference that covers the pixel $j$. Therefore, the final forward change level of this specific pixel $j$ is the mean of this set as

$$DI_j^{\mathbf{Y}} = \frac{1}{|F_j^{\mathbf{Y}}|} \sum_{f_i^{\mathbf{Y}} \in F_j^{\mathbf{Y}}} f_i^{\mathbf{Y}} \quad (7)$$

At the same time, the backward $DI^{\mathbf{X}}$ can also be generated in a similar way as the $DI^{\mathbf{Y}}$. Then, we can obtain the final $DI^{final}$ by fusing the forward $DI^{\mathbf{Y}}$ and backward $DI^{\mathbf{X}}$ as

$$DI^{final} = DI^X / mean(DI^X) + DI^Y / mean(DI^Y) \quad (8)$$

Algorithm 2 listed in Table II summarizes the procedure of generating the ALSC based DI. Given the set $\Lambda$ of all target patches spaced by a patch step size $\Delta_p \in [1, 2p+1]$, the number of target patches can be reduced by $\Delta_p^2$, which can accelerate the algorithm.

TABLE II
THE GENERATION STEPS OF ALSC BASED DI.

| Algorithm 2. ALSC based DI |
|---|
| **Input:** Images of $\mathbf{X}$ and $\mathbf{Y}$, parameters of $p$, $\omega$, $\Delta_p$, $\Delta_s$ and $K$. |
| 1. Calculation of the forward and backward structure differences<br>  **for** all the target patches $\mathbf{X}_i$, $\mathbf{Y}_i$, $i \in \Lambda$ **do**<br>    Compute the distances $dist_i^{\mathcal{X}}$ and $dist_i^{\mathcal{Y}}$ with different criteria.<br>    Compute the similarities $S_i^{\mathbf{X}}$ and $S_i^{\mathbf{Y}}$ by using Algorithm 1.<br>    Compute the structure difference $f_i^{\mathbf{X}}$ and $f_i^{\mathbf{Y}}$.<br>    Add $f_i^{\mathbf{X}}$ and $f_i^{\mathbf{Y}}$ to the sets $F_j^{\mathbf{X}}$ and $F_j^{\mathbf{Y}}$, respectively.<br>  **end for** |
| 2. Computation of the forward and backward difference images<br>  **for** all the pixels $j$, $j = 1, \cdots, MN$ **do**<br>    Compute the $DI^{\mathbf{Y}}$ and $DI^{\mathbf{X}}$.<br>  **end for** |
| 3. Fusion of the forward and backward difference images<br>  $DI^{final} = DI^X / mean(DI^X) + DI^Y / mean(DI^Y)$ |

*D. Obtain the binary change map*

The final CM can be calculated by solving the binary segmentation problem after obtaining the fused DI. Then, the thresholding method (such as the Otsu threshold method [18]) or the clustering method (such as the PCAKM [19] that using the principal component analysis and the $k$-means) can be utilized to obtain the binary CM.

III. EXPERIMENTAL RESULTS

*A. Data description*

The performance of the ALSC based heterogeneous CD method is evaluated on four pairs of heterogeneous data sets, as shown in Figs. 2(a)-(c).

(1) Sardinia data set: a pair of $300 \times 412$ near-infrared (NIR) band/optical images. The NIR image is sensed by Landsat-5, and the optical image is obtained from Google Earth with three bands. The ground truth shows the Lake expansion on Sardinia, Italy. (2) Shuguang data set: a pair of $593 \times 921$ SAR/optical images. The SAR image is sensed by Radarsat-2, and the optical image is obtained from Google Earth with three bands. The ground truth shows the land use changes of Shuguang Village, Shandong, China. (3) Wuhan data set: a pair of $495 \times 503$ SAR/optical images. The SAR image is sensed by Radarsat-2, and the optical image is obtained from Google Earth with three bands. The ground truth shows the new roads and buildings in Wuhan City, China. (4) California data set: a pair of $875 \times 500$ multispectral/SAR images. The multispectral image is sensed by Landsat-8 with 11 bands, and the SAR image is sensed by Sentinel-1A with three channels (two channels are VV and VH polarization data, and the third channel is the ratio between them). The ground truth shows a flood in California, USA, which is constructed by Luppino et al. [13].

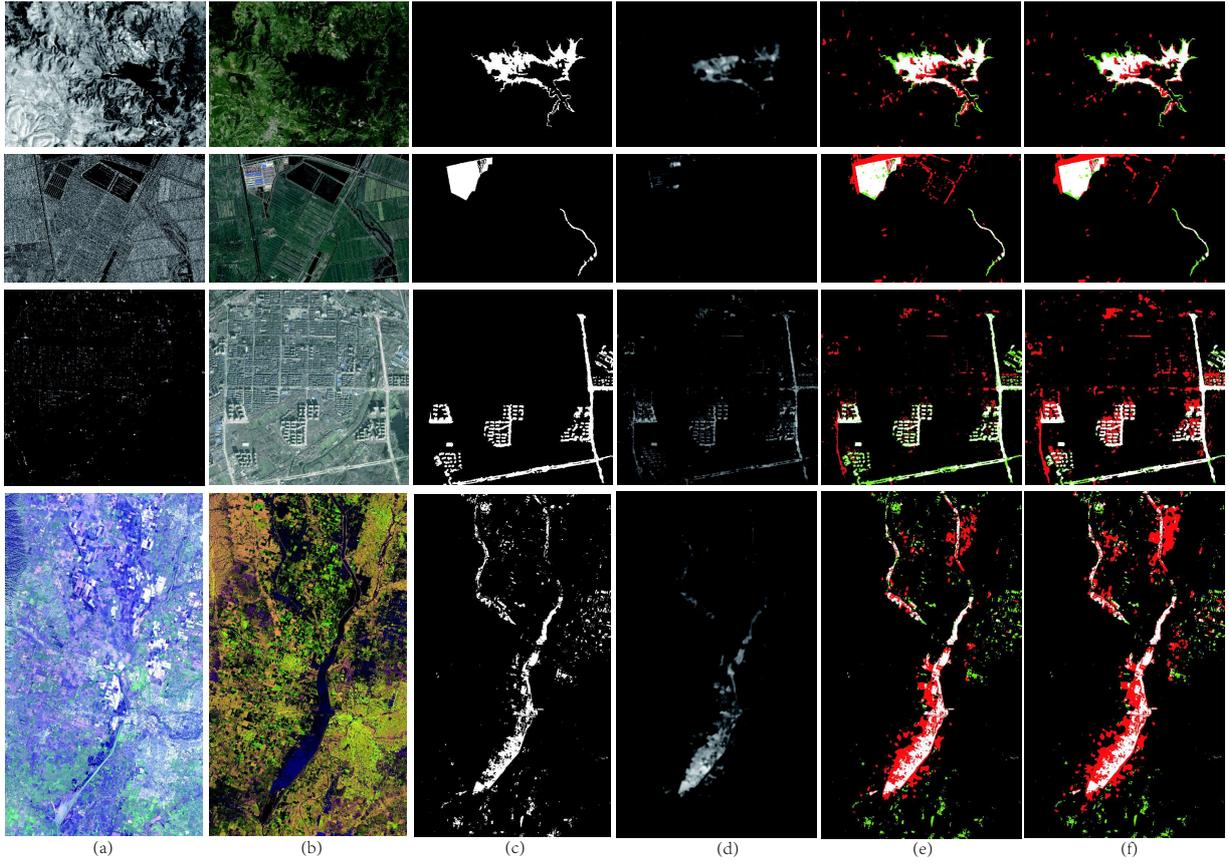

Fig. 2. ALSC based DI and binary CM on different data sets. From top to bottom, they correspond to Sardinia, Shuguang, Wuhan and California data sets, respectively. From left to right are: (a) pre-event image; (b) post-event image; (c) the ground truth image; (d) ALSC based DI; (e) binary CM of ALSC-O; (f) binary CM of ALSC-P. In the binary CM, White: true positives (TP); Red: false positives (FP); Black: true negatives (TN); Green: false negatives (FN).

### B. Experimental results and analysis

As listed in Table II, the parameters of ALSC based method are the patch size $p$, the patch step size $\Delta_p$, the search window size $\omega$, the search step size $\Delta_s$, and the number of neighbors $K$. For all the experimental results, we fix $\omega = 75p$, $\Delta_p = p$, $\Delta_s = 2p+1$, $K = 35$, and vary the $p$ from 1 to 4 to select the best one for each data set. Specifically, we set $p = 1$ for Wuhan, $p = 2$ for Sardinia and California, and $p = 3$ for Shuguang.

Fig. 2(d) shows the ALSC generated DI of different data sets. It can be found that the proposed ALSC can well build the relationship between heterogeneous images, which can highlight changes in DI. Fig. 3 plots the empirical receiver operating characteristics (ROC) curves of ALSC generated DI, and shows that the quality of these DI is very high, which gain a large area under the ROC curves as 0.970, 0.979, 0.957 and 0.925 for Sardinia, Shuguang, Wuhan and California data sets, respectively.

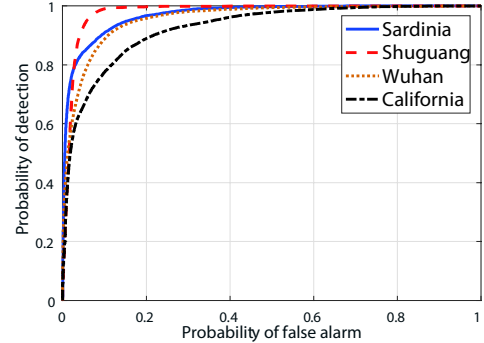

Fig. 3. ROC curves of ALSC generated DI.

With these ALSC based DI, we can obtain the final CM by using the Otsu thresholding (denoted as ALSC-O for short) and PCAKM (denoted as ALSC-P for short), as shown in Figs. 2(e)-(f). Four measures are considered as the evaluation criteria of CM: the false positives (FP) rate, the false negatives (FN) rate, the overall accuracy (OA), and the Kappa coefficient (KC). Table III reports the quantitative measures of ALSC based binary CM of different heterogeneous data sets. From Figs. 2(e)-(f) and Table III, we can see that the changed and unchanged area are well detected with relatively small FN and FP.

TABLE III
QUANTITATIVE MEASURES OF ALSC BASED BINARY CM.

| Data sets | ALSC-O | | | ALSC-P | | |
|---|---|---|---|---|---|---|
| | FP(%) | FN(%) | KC | FP(%) | FN(%) | KC |
| Sardinia | 2.42 | 1.35 | 0.6983 | 2.10 | 1.40 | 0.7134 |
| Shuguang | 3.65 | 0.52 | 0.6410 | 3.11 | 0.55 | 0.6693 |
| Wuhan | 1.76 | 2.83 | 0.5950 | 5.82 | 1.01 | 0.5849 |
| California | 4.07 | 1.57 | 0.4697 | 6.05 | 1.31 | 0.4201 |

TABLE IV
ACCURACY RATE OF CM GENERATED BY DIFFERENT METHODS ON DIFFERENT DATA SETS (THE OPTIMAL TWO VALUES OF EACH DATA SET ARE WRITTEN IN BOLD).

| Sardinia | OA | Shuguang | OA | Wuhan | OA | California | OA |
|---|---|---|---|---|---|---|---|
| ALSC-O | 0.962 | ALSC-O | 0.958 | ALSC-O | **0.954** | ALSC-O | **0.944** |
| ALSC-P | **0.965** | ALSC-P | 0.963 | ALSC-P | 0.932 | ALSC-P | 0.926 |
| FP-MS[14] | 0.928 | FP-MS[14] | 0.942 | LTFL[8] | 0.952 | FP-MS[14] | **0.952** |
| PSGM[3] | 0.961 | PSGM[3] | **0.977** | PSGM[3] | 0.953 | AM-IR[13] | 0.933 |
| M3CD[11] | **0.964** | NPSG[15] | 0.975 | NPSG[15] | **0.958** | NPSG[15] | 0.941 |
| RMN[20] | 0.847 | RMN[20] | 0.884 | | | | |
| AFL-DSR[21] | 0.929 | AFL-DSR[21] | **0.980** | | | | |

In order to further evaluate the performance of ALSC, some representative and state-of-the-art (SOTA) methods including PSGM [3], LTFL [8], M3CD [11], AM-IR [13], fractal projection and Markovian segmentation-based method (FP-MS) [14], nonlocal patch similarly graph-based method (NPSG) [15], reliable mixed-norm-based heterogeneous CD method (RMN) [20], and anomaly feature learning based deep sparse residual model (AFL-DSR) [21], are selected for comparison as summarized in Table IV. We can see that ALSC can achieve better or quite competitive performance by comparing with these SOTA methods, and ALSC shows the ability to gain consistent good results on different data sets. The average accuracy rates obtained on four heterogeneous data set of the ALSC with Otsu thresholding and PCAKM are 95.46% and 94.66%, respectively.

## IV. CONCLUSION

In this letter, we focus on the problem of change detection in heterogeneous remote sensing. We explore the ALSC to establish a relationship between heterogeneous images, which exploits the inherent structure property of images and appeals quite imaging modality-invariant. The core idea of the ALSC is that the local structure of each target patch in one image will keep invariance in the other image when there is no change occurred; on the contrary, once a change occurred within this target patch, this local invariance will no longer be maintained. Therefore, the ALSC based CD method can be divided into two process. Firstly, the adaptive local structures are constructed for each input image. Then, the structure differences are compared in the same image domain by graph projection, which can avoid the heterogeneous data confusion. Experimental results clearly show that the ALSC based CD method can achieve effective performance on different heterogeneous data sets.